\documentclass{article}

\usepackage{microtype}
\usepackage{graphicx}
\usepackage{subfigure}
\usepackage{booktabs} 
\usepackage{pifont}
\usepackage{amsfonts}
\usepackage{hyperref}
\usepackage{natbib}   
\usepackage{amsmath}
\usepackage{graphicx}
\usepackage{array}
\usepackage{enumitem}
\usepackage{algorithm}
\usepackage{algorithmic}

\usepackage[accepted]{icml2021}

\icmltitlerunning{Pesti-Gen by Taehan Kim and Wonduk Seo}

\begin{document}

\twocolumn[

\icmltitle{Pesti-Gen: Unleashing a Generative Molecule Approach \\for Toxicity Aware Pesticide Design}

\icmlsetsymbol{equal}{*}

\begin{center}
    \begin{tabular}{cc}
        \begin{tabular}{c}
            \textbf{Taehan Kim\textsuperscript{*}\textsuperscript{†}} \\
            University of California, Berkeley, USA \\
            \texttt{terry.kim@berkeley.edu}
        \end{tabular}
        &
        \begin{tabular}{c}
            \textbf{Wonduk Seo\textsuperscript{*}} \\
            Peking University, China \\
            \texttt{seowonduk@pku.edu.cn}
        \end{tabular}
    \end{tabular}

\end{center}

\vskip 0.3in
]
\footnotetext[1]{\textsuperscript{*}Taehan and Wonduk contributed equally to this work.}
\footnotetext[2]{\textsuperscript{†}Taehan is the corresponding author for this work.}

\begin{abstract}
Global climate change has reduced crop resilience and pesticide efficacy, making reliance on synthetic pesticides inevitable, even though their widespread use poses significant health and environmental risks. While these pesticides remain a key tool in pest management, previous machine-learning applications in pesticide and agriculture have focused on classification or regression, leaving the fundamental challenge of generating new molecular structures or designing novel candidates unaddressed. In this paper, we propose \textit{\textbf{Pesti-Gen}, a novel generative model based on variational auto-encoders, designed to create pesticide candidates with optimized properties for the first time}. Specifically, \emph{Pesti-Gen} leverages a two-stage learning process: an initial pre-training phase that captures a generalized chemical structure representation, followed by a fine-tuning stage that incorporates toxicity-specific information. The model simultaneously optimizes over multiple toxicity metrics, such as (1) livestock toxicity and (2) aqua toxicity to generate environmentally friendly pesticide candidates. Notably, \emph{Pesti-Gen} achieves approximately 68\% structural validity in generating new molecular structures, demonstrating the model's effectiveness in producing optimized and feasible pesticide candidates, thereby providing a new way for safer and more sustainable pest management solutions.
\end{abstract}

\section{Introduction} 
\label{submission}
Pesticides have been pivotal in global food security and agricultural productivity since their widespread adoption in the mid-20th century. Their importance was underscored by Paul Hermann Müller’s discovery of DDT’s insecticidal properties, a breakthrough that revolutionized pest control \cite{muller1948}. Even today, pesticides remain a cornerstone in reducing crop losses. With estimates suggesting that, in their absence, key staple crops could experience up to a 30\% yield loss due to pests and pathogens, translating into hundreds of billions of dollars in global economic losses \cite{rizzo2021plant}. Despite their undeniable benefits, the quest to develop safe and effective pesticides continues to face significant hurdles. Cross-resistance, toxicity, and unintended ecological consequences persist as major concerns \cite{araujo2023buzz}. These challenges are further highlighted by regulatory bans—such as those on imidacloprid and clothianidin between 2016 and 2018—and by findings from the European Food Safety Authority (EFSA), which identified neonicotinoids as posing substantial risks to pollinators, including both wild bees and honeybees \cite{araujo2023buzz}.

In addition to direct toxicity, the shifting landscape of global climate change exacerbates the need for eco-friendly pesticide solutions. Rising temperatures not only affect pest behavior but also reduce the effectiveness and persistence of many pesticides \cite{hannigan2023effects, iltis2022warming, rhodes2020analysis}. Seminal studies demonstrate that increased temperatures can diminish a chemical’s residual activity, often necessitating more frequent reapplication, thereby raising both economic and environmental costs~\cite{lichtenstein1959persistence, walker1983prediction, nokes1992predicting, garcia1994persistence, ahmad2003dissipation, bailey2004climate}. Although machine learning has been widely used in pesticide and agricultural research—primarily for classification and regression tasks, such as predicting toxicity or categorizing pesticides~\cite{anwar2023exploring}—these approaches have not yet tackled the critical need to generate novel molecular structures or design new candidates.

To address the challenge of developing safer and more sustainable pesticides, we propose a novel generative framework, \textit{Pesti-Gen}. Specifically, our approach employs a two-stage learning process: first, pre-training on a large corpus of general molecular structures to establish a comprehensive latent space capturing broad chemical features; and second, fine-tuning on specialized toxicity metrics, including livestock toxicity (\textit{LD50}) and aquatic eco-toxicity (\textit{LC50}). These metrics are meticulously curated based on WHO toxicity classification standards and concentration thresholds outlined by \citet{rda_aqua_ecotoxicity}. To enable meaningful evaluations, we curated a custom dataset combining detailed toxicity grades and SMILES representations of diverse pesticide candidates, addressing the lack of publicly available benchmarks for generative pesticide design.

Furthermore, to validate \textit{Pesti-Gen}, we conducted comprehensive experiments encompassing chemical validity assessments, physicochemical property comparisons (e.g., LogP and SAS scores), and structure plausibility through SMILES validity checks. Our framework achieved approximately \textbf{68\% chemical validity} in generating pesticide-like candidates with optimized toxicity levels. Additionally, our results demonstrated the model's capability to balance hydrophobicity, synthetic accessibility, and toxicity reduction, underscoring its practicality for real-world applications. By integrating cutting-edge molecular generative techniques with eco-toxicity constraints, \textit{Pesti-Gen} lays a robust foundation for the development of innovative, environmentally conscious pesticide solutions.

\begin{table*}[t]
\centering
\small
\renewcommand{\arraystretch}{1.2}
\resizebox{\textwidth}{!}{
\begin{tabular}{l l c c c}
\toprule
\textbf{Class  \newline(Livestock Toxicity)} 
& \textbf{Description} 
& \textbf{LD\textsubscript{50} (Oral, rat, mg/kg BW)} 
& \textbf{LD\textsubscript{50} (Dermal, rat, mg/kg BW)} 
& \textbf{Mapped Scale (Our Data)} \\
\midrule
\textbf{Ia} 
& Extremely hazardous 
& $< 5$ 
& $< 50$ 
& 1000 \\
\textbf{Ib} 
& Highly hazardous 
& 5--50 
& 50--200 
& 1000 \\
\textbf{II} 
& Moderately hazardous 
& 50--2000 
& 200--2000 
& 100 \\
\textbf{III} 
& Slightly hazardous 
& $>2000$ 
& $>2000$ 
& 10 \\
\textbf{IV} 
& Unlikely to present acute hazard 
& $\ge 5000$ 
& $\ge 5000$ 
& 0 \\
\bottomrule
\end{tabular}
}
\caption{\textbf{Livestock Toxicity classification based on WHO guidelines for pesticide classification based on LD\textsubscript{50} values, alongside our mapped scale before normalization.} The mapped scale assigns 10 to values in the range 5–50, 100 to values in the range 50–2000, and 1000 to values above 2000, following the guided ranges shown in the table.}
\label{table:who_guideline}
\vspace{-1em}
\end{table*}

\begin{table*}[t]
\centering
\small
\renewcommand{\arraystretch}{1.2}
\begin{tabular}{p{2.5cm} p{5.5cm} p{3.5cm} p{3cm}}
\toprule
\textbf{Class \newline (Aquatic Toxicity)} 
& \textbf{Description} 
& \textbf{LC\textsubscript{50} (mg/L, 48 hours)} 
& \textbf{Mapped Scale (Our Data)} \\
\midrule
\textbf{I} 
& Extremely toxic to aquatic organisms 
& $< 0.5$ 
& 16 \\
\textbf{II} 
& Highly toxic to aquatic organisms 
& 0.5--2.0 
& 4 \\
\textbf{II s} 
& Moderately toxic to aquatic organisms (subclass) 
& $\text{Not measurable, } < 0.1$ 
& 4 \\
\textbf{III} 
& Slightly toxic to aquatic organisms 
& $> 2.0$ 
& 1 \\
\textbf{Exempt} 
& Unlikely to cause harm to aquatic organisms 
& -- 
& 0 \\
\bottomrule
\end{tabular}
\caption{\textbf{Aquatic Toxicity classification for pesticides based on aquatic LC\textsubscript{50} values and mapped scale.} The table describes toxicity levels, corresponding LC\textsubscript{50} (lethal concentration for 50\% of the test population over 48 hours), and the mapped numerical scale used in our dataset. Class II and Class II S are mapped to the same number because it is specific to the Korean ecosystem.}
\label{table:aquatic_toxicity}
\vspace{-1em}
\end{table*}

\begin{table*}[t]
\centering
\small
\renewcommand{\arraystretch}{1.2} 
\resizebox{\textwidth}{!}{ 
\begin{tabular}{p{5.3cm} p{3cm} p{2.6cm} p{2cm} p{2cm} p{6.7cm}}
\toprule
\textbf{\small AGCHM\_ENG\_NM} & \textbf{\small Livestock Toxicity \newline (LD 50)} & \textbf{\small Aqua Ecotoxicity \newline (LC 50)} 
& \textbf{\small Livestock Tox.\ Metrics} & \textbf{\small Aqua Tox.\ Metrics} 
& \textbf{\small SMILES (Abbrev.)} \\
\midrule
\footnotesize Benzobicyclon + Pyrazosulfuron-ethyl + Pyriminobac-methyl 
& \footnotesize Class IV 
& \footnotesize Class III 
& \footnotesize 0.01 
& \footnotesize 0.0625 
& \small \texttt{CS(=O)(=O)C1=CC(=C(C=C1)C(=O)C2=C...} \\
\midrule
\footnotesize Fenquinotrione + Imazosulfuron + Pyriminobac-methyl 
& \footnotesize Class IV 
& \footnotesize Class III 
& \footnotesize 0.01 
& \footnotesize 0.0625 
& \small \texttt{COC1=CC=C(C=C1)N2C3=C(C(=CC=C3)Cl...} \\
\midrule
\footnotesize Fentrazamide + Metazosulfuron 
& \footnotesize Class IV 
& \footnotesize Class III 
& \footnotesize 0.01 
& \footnotesize 0.0625 
& \small \texttt{CCN(C1CCCCC1)C(=O)N2C(=O)N(N=N2)...} \\
\midrule
\footnotesize Fenoxasulfone + Fenquinotrione + Imazosulfuron 
& \footnotesize Class IV 
& \footnotesize Class III 
& \footnotesize 0.01 
& \footnotesize 0.0625 
& \small \texttt{CCOC1=C(C=C(C(=C1)Cl)CS(=O)(=O)C2...} \\
\bottomrule
\end{tabular}
}
\caption{
\textbf{Sample of curated data from the Gyeonggi Province Pesticide Dataset 
(Abbreviated SMILES and Metrics).} Two separate toxicity metrics (livestock and aqua) was compiled. Each Tox.\ metrics were normalized based on the class, and converted to float metrics. SMILES strings were obtained using PubChemPy.
}
\label{table:sample_data}
\vspace{-1em}
\end{table*}

\section{Related Work}
Machine learning applications in agriculture have traditionally focused on classification and detection tasks, particularly in computer vision-driven research such as plant disease identification or pest recognition \cite{anwar2023exploring}. These approaches typically aim to enhance early detection and management strategies but often do not address the inherent toxicity or environmental impact of pesticides. Beyond image-based tasks, other machine learning endeavors center on pesticide classification and property prediction using physicochemical or dissipation-related parameters. For instance, Shen et al. \cite{shen2022predicting} leveraged previously reported pesticide data to perform classification based on half-life, which is a vital measure influencing both toxicity levels and recommended dosages. Similarly, Quantitative Structure-Activity Relationship (QSAR) techniques \cite{isarankura2009practical} and Structure Alerts (SA) models have been employed to predict toxicity, filtering out chemically active structures likely to exhibit adverse effects. Ensemble-based methods have further improved the accuracy and robustness of toxicity predictions \cite{kwon2019comprehensive}. While these predictive and classification models provide valuable tools for optimizing the use of existing pesticides, they fall short in enabling the discovery of novel, eco-friendly compounds.

In contrast, the drug discovery domain has seen significant advancements in methods for generating new small molecules with desired properties, demonstrating the potential of data-driven approaches to produce viable chemical candidates. Unlike predictive frameworks that estimate properties of known molecules, generative modeling tools are designed to create entirely new molecular structures. For example, ChemicalVAE \cite{gomez2018automatic} exemplifies such an approach, utilizing a Variational Autoencoder (VAE) to learn latent representations of molecules and to generate candidate compounds optimized for specific properties. Further advancing these ideas, Reinvent4 \cite{loeffler2024reinvent}, an open-source molecular design platform released by AstraZeneca, demonstrates how industrial collaborators are embracing machine learning to expedite the drug discovery pipeline. Recent advancements have also extended generative methodologies into the 3D domain, where small molecules are represented as voxelized structures, and new candidates are generated by perturbing these voxel grids \cite{nowara2024nebula}. 

Despite these advances, adapting generative methods to develop ecologically safer pesticides remains an open challenge. Although computational strategies from drug discovery and pesticide classification share common ground as molecular discovery tasks, they differ significantly in how toxicity is measured and prioritized, especially given the broader ecological considerations inherent to agriculture. Furthermore, there is a noticeable lack of foundational models specifically tailored for molecule generation in the context of pesticide design, underscoring a critical gap in this domain. Thus, this work aims to address this gap by bridging the divide between existing computational methods and their application in agriculture, particularly in pesticide design.


\section{Datasets}

As no publicly available benchmarks or established small-molecule datasets specifically designed for novel pesticide candidate generation exist, we compiled a custom dataset tailored to this task. While datasets focusing on metrics such as half-life are available, our dataset uniquely incorporates both livestock toxicity scores (based on \textit{LD\textsubscript{50}}) and aqua toxicity scores (based on \textit{LC\textsubscript{50}}). This dataset includes the molecular structures of relevant pesticides in SMILES format alongside their corresponding toxicity scores, providing a resource for training and evaluating our model under practical constraints.

\subsection{Limitation of Initial Half-Life Based Data}
A common metrics used to evaluate the toxicity of pesticides and environmental risk indicators is \emph{half-life} (the time required for a pesticide to degrade by half).
Initially, the training dataset for \emph{Pesti-Gen} was curated by focusing on half-life based on previous works, specifically by modifying the dataset presented in the work of Shen et al.~\cite{shen2022predicting} which compiled the historically compiled data \cite{fantke2013variability} and pesticide candidates. Small molecules were converted to SMILE representation and collected based on the pesticide chemical names, paired with the half-life as a toxicity feature. 

However, half-life is a volatile metrics that is highly dependent on other environment factors when measured. Referring to the work of Fantke and Juraske, the log-linear model for predicting accurate half-lives (Model III) for pesticide dissipation in plants is modeled by a equation that takes in temperature effects ($\beta'^{\circ}_{T}$), molecular weight ($\beta'^{\circ}_{\text{MW}}$), saturation vapor pressure ($\beta'^{\circ}_{P_{V}}$), and octanol partition coefficient ($\beta'^{\circ}_{K_{\text{OW}}}$) as a parameter \cite{fantke2014estimating}. NAFTA emphasizes the necessity of keeping variables such as soil type, temperature, moisture, and soil incubation prior to pesticide application constant \cite{spatz2015standard}. In other words, it is a highly variable metrics that needs careful control and normalization across different reports.

There are other variables such as temperature and location (differing soil component and weather) other than the actual toxicity or pesticide chemical component that can contribute to the variability of half-life being reported. In result, Shen et al. \cite{shen2022predicting} focused on predicting the intervals that pesticide dissipation would fall into rather than directly predicting its value. In other words, half-life is not a suitable metrics to be directly optimized after. The curated dataset focusing on half-life as an optimization metrics was discarded due to the high variability of half-life to be directly used as a metric. Additionally, the dataset, which was based on the work of \cite{fantke2013variability}, contained limited number of 4513 data points and 311 unique pesticides \cite{shen2022predicting}.

\subsection{Custom Dataset Based on \(LD_{50}\) and \(LC_{50}\) Toxicity}

As an alternative to resolve this variability, final training pesticide dataset for \textit{Pesti-Gen} was compiled based on \ding{172} \textit{LD\textsubscript{50}} and \ding{173} \textit{LC\textsubscript{50}}, as these metrics introduce less variability compared to environmental factors. \(LD\textsubscript{50}\) (Lethal dose) refers to the dose required to kill 50\% of a population and can be mathematically expressed as:
\begin{equation}
LD_{50} = \frac{X}{W}
\end{equation}
where \(LD\textsubscript{50}\) is the lethal dose for 50\% of the population, \(X\) is the amount of substance administered (in mg), and \(W\) is the body weight of the organism (in kg) \cite{damalas2011pesticide}. This \(LD\textsubscript{50}\) metrics assesses risks to mammals and livestock. In addition to \(LD\textsubscript{50}\), the dataset incorporates \(LC\textsubscript{50}\) (Lethal Concentration 50), which measures the concentration of a substance required to kill 50\% of a test population in aquatic environments. It is expressed in units of mg/L over a specified exposure duration, typically 48 hours:
\begin{equation}
LC_{50} = \frac{M}{V}
\end{equation}
where \(LC\textsubscript{50}\) is the lethal concentration for 50\% of the population, \(M\) is the mass of the pesticide (in mg), and \(V\) is the volume of the solution (in L). \textbf{\(LC_{50}\)} and Aquatic Ecosystem Relevance: As outlined in the aqua ecotoxicity classifications \citet{rda_aqua_ecotoxicity}, the \(LC_{50}\) metric provides a critical understanding of how pesticides affect aquatic ecosystems. This measure evaluates the toxicity of a pesticide to aquatic organisms such as fish, crustaceans, and algae over a defined exposure period (typically 48 hours). The classification of \(LC_{50}\) values follows established toxicity thresholds to distinguish levels of environmental risk.

The combination of \(LD\textsubscript{50}\) for livestock and \(LC\textsubscript{50}\) for aquatic ecosystems provides a comprehensive evaluation of pesticide toxicity. While \(LD\textsubscript{50}\) assesses risks to mammals and livestock, \(LC\textsubscript{50}\) emphasizes environmental impact on aquatic ecosystems. Together, these metrics enable a balanced approach to pesticide safety, addressing both livestock and aqua ecosystem. The ranges for these toxicity metrics are further detailed in Table~\ref{table:who_guideline} for \(LD\textsubscript{50}\) and Table~\ref{table:aquatic_toxicity} for \(LC\textsubscript{50}\).

\subsection{Gyeonggi Province Pesticide Open Dataset}
\label{subsec:gyeonggi}
To curate a pesticide dataset that accurately accounts the \textit{LD50} metrics as toxicity measure, dataset from Gyeonggi Province in Korea was   referred\footnote{\href{https://data.gg.go.kr/portal/data/service/selectServicePage.do?\&infId=XM0L7AK0UL00KYARRMU929751298&infSeq=1\#none}{Gyeonggi Province Pesticide Open Dataset}}. The dataset was translated, and curated due to its valuable features. It provides detailed human/livestock toxicity and aqua ecotoxicity grades based on \textit{LD50}, with the human/livestock toxicity classifications aligned with the WHO (Table~\ref{table:who_guideline}) guidelines, ensuring reliable and standardized risk assessments. 

Additionally, the dataset provides geographic coordinates (latitude and longitude) for each entry, which are crucial for agriculture-related studies where environmental factors such as soil and temperature significantly influence pesticide behavior. Moreover, the dataset also includes and offers a diverse pesticide formed in mixture form (e.g. Benzobicyclon+Pyrazosulfuron-ethyl+Pyriminobac-methyl), including unique combinations, adding robustness and depth to the analysis. These attributes make it a highly compatible for the purpose of our \emph{Pesti-Gen} model, obtaining stable metrics paired with diverse pesticide for training. In detail, the curated custom dataset comprises $56,360$ data points and $520$ unique pesticides, significantly improving upon previous datasets, which included only 4513 datapoints and $311$ unique pesticides.

\subsection{Custom Dataset: \ding{172} Livestock \ding{173} Aqua Ecotoxicity}

Relevant toxicity classification labels provided by the Gyeonggi Province Pesticide Open Dataset were converted to numerical values for model training. Matching the guideline range scale, the categorical classification metrics were mapped to numerical scales (\emph{e.g., 10 for livestock toxicity and 4 for aquatic toxicity}) and subsequently normalized to a scale of 0 to 1. For instance, pesticides classified as Class II for livestock toxicity were mapped to 100 and normalized accordingly, while aquatic toxicity classifications such as Class II and Class II S are mapped to the same number because Class II S is specific to the Korean ecosystem. The specific mapping logic applied is as follows:  
\begin{itemize} 
    \item For livestock toxicity:  
    \begin{itemize} 
        \item \textbf{High Toxicity (Class II)}: Mapped to 1000 
        \item \textbf{Moderate Toxicity (Class III)}: Mapped to 100 
        \item \textbf{Low Toxicity (Class IV)}: Mapped to 10 
        \item \textbf{Unclassified}: Mapped to 0 
    \end{itemize} 
    \item For aquatic toxicity:  
    \begin{itemize} 
        \item \textbf{Class I}: Mapped to 16 (highest risk to aquatic ecosystems) 
        \item \textbf{Class II and Class II S}: Mapped to 4 (to account for similar aquatic impact) 
        \item \textbf{Class III}: Mapped to 1 (lowest risk) 
        \item \textbf{Exempt}: Mapped to 0 
    \end{itemize} 
\end{itemize}

After applying these mappings, the numerical values were normalized using the formula:  
\begin{equation} 
\text{Normalized Value} = \frac{\text{Value} - \text{Min Value}}{\text{Max Value} - \text{Min Value}} 
\end{equation} 

This process ensures that higher toxicity values remain proportional across both livestock and aquatic toxicity metrics, making them suitable for training a machine learning model. Chemical names from the dataset were also mapped to their respective SMILES strings, with mixtures denoted using a period (`.'). A sample of the curated data after this normalization and mapping process is presented in Table 3.

In addition to livestock toxicity grades, the aquatic toxicity grades explicitly account for the unique toxicity risks to marine ecosystems. These grades are classified into four categories:  
\begin{itemize} 
    \item \textbf{Class I Pesticides}: Should not be used in rice paddy fields near fish farms, reservoirs, water supply sources, or marine areas due to the high risk of runoff. Even non-rice paddy fields should avoid these pesticides in areas where wind or rain could cause contamination of water bodies.  
    \item \textbf{Class II Pesticides}: Broad application should be avoided near water sources, particularly fields close to fish farms, reservoirs, or marine environments, to prevent contamination from wind drift or rain.  
    \item \textbf{Class II S Pesticides}: Similar precautions as Class II, with additional emphasis on avoiding broad application near sensitive aquatic environments.  
    \item \textbf{Class III Pesticides}: Should not be used in rice paddy fields within designated protected water source areas.  
\end{itemize}

This distinction is crucial, as some pesticides may pose limited harm to livestock but cause significant damage to aquatic ecosystems. The mapping logic ensures that both livestock and aquatic toxicity risks are appropriately weighted, allowing the model to account for the diverse environmental impacts of pesticide use.

\section{Methodology}

\begin{figure*}[!htpb]
    \centering
    \includegraphics[width=0.8\linewidth]{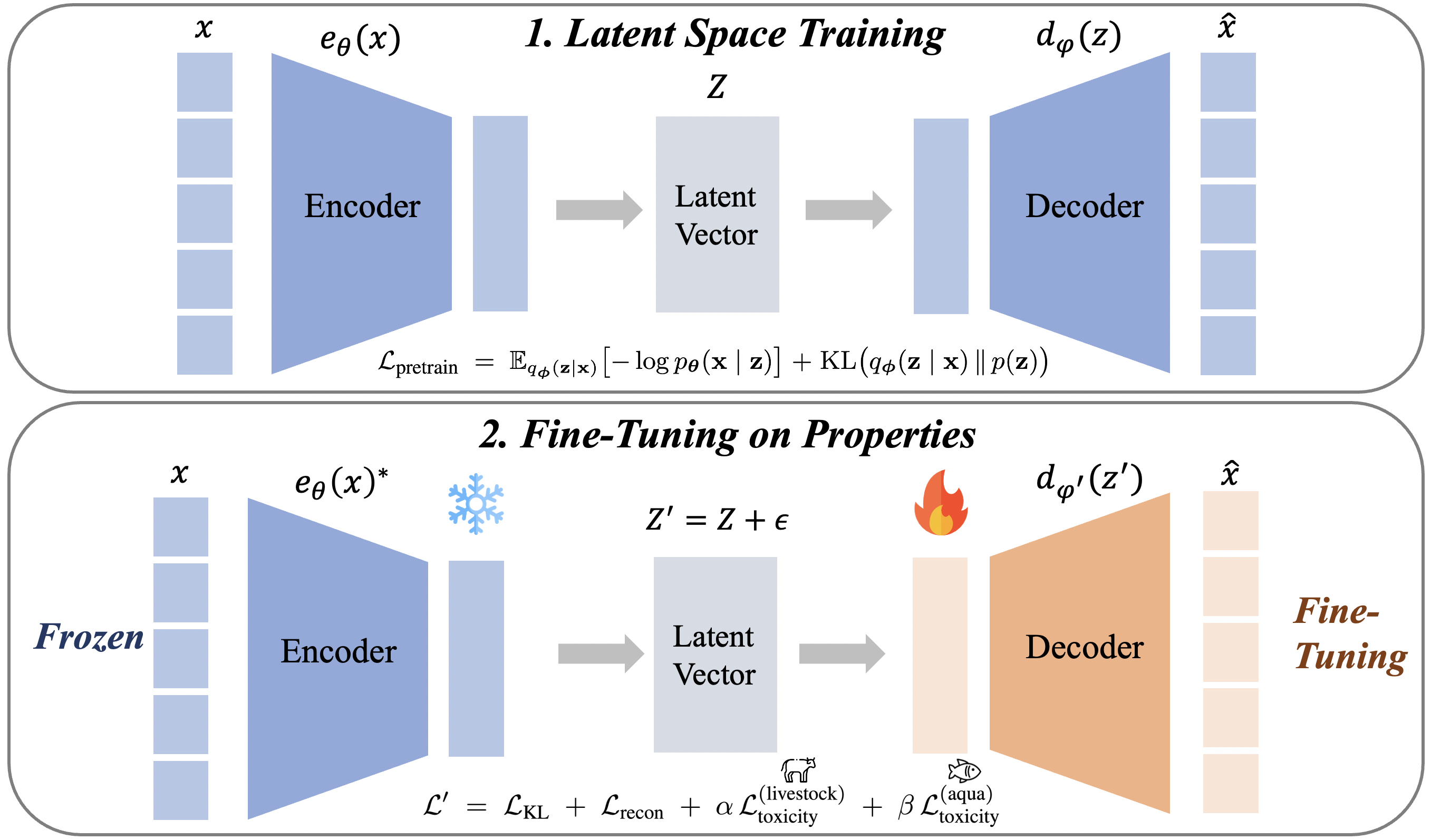}
    \caption{\textbf{Overview of the two-stage Pesti-Gen training process}: 
(1) \emph{Latent Space Training} learns a general molecular representation using reconstruction and KL loss; 
(2) \emph{Fine-Tuning on Properties} incorporates toxicity-specific properties (\(\textit{LD}_{50}\) and ecological toxicity) with targeted loss components to optimize pesticide candidates.
}
    \label{fig:framework}
\end{figure*}

In this section, we describe our two-stage training procedure for \emph{Pesti-Gen}, which aims to capture both general molecular properties and specific pesticide toxicity measures. (1) During the first stage, the model learns a broad latent representation of SMILES structures; (2) in the second stage, the learned representations are refined to explicitly incorporate toxicity constraints. An overview of the  framework is provided in Figure~\ref{fig:framework}.

\subsection{Latent Space Training with VAE}
\label{subsec:stage1}

Our \emph{Pesti-Gen} model initially follows the architecture outlined by G{\'o}mez et al.~\cite{gomez2018automatic}, using a Variational Autoencoder (VAE) \cite{kingma2013auto} trained on a large corpus of SMILES strings based on the ZINC dataset \cite{irwin2005zinc}. The dataset consists of 50{,}000 diverse molecular structures, all of which were specifically utilized for training VAE model\footnote{\url{https://github.com/aksub99/molecular-vae/tree/master/data}}. We trained the model for 30 epochs with a learning rate of \(5 \times 10^{-5}\). The primary objectives in this stage are: \ding{172} \textbf{SMILES Reconstruction:} Learning to accurately encode and decode molecular representations; \ding{173} \textbf{Latent Space Structuring:} Creating a continuous and meaningful latent embedding that captures key chemical features.

The training process simultaneously minimizes both the reconstruction loss and the KL divergence, as defined by the following loss function:
\begin{equation}
\small
\mathcal{L}_{\text{pretrain}} \;=\; \mathbb{E}_{q_{\boldsymbol{\phi}}(\mathbf{z} \mid \mathbf{x})} \bigl[-\log p_{\boldsymbol{\theta}}(\mathbf{x} \mid \mathbf{z})\bigr]
\;+\;
\mathrm{KL} \bigl(q_{\boldsymbol{\phi}}(\mathbf{z} \mid \mathbf{x}) \,\|\, p(\mathbf{z})\bigr),
\end{equation}
where \(q_{\boldsymbol{\phi}}(\mathbf{z} \mid \mathbf{x})\) represents the encoder modeling the variational posterior distribution, \(p_{\boldsymbol{\theta}}(\mathbf{x} \mid \mathbf{z})\) is the decoder modeling the reconstruction likelihood, and \(p(\mathbf{z})\) is the prior distribution over latent variables, typically assumed to be \(\mathcal{N}(\mathbf{0}, \mathbf{I})\). 

The reconstruction loss encourages accurate decoding of input SMILES from latent representations to ensure chemical plausibility, while the KL divergence regularizes the latent space by aligning it with the prior distribution, ensuring smoothness and meaningful sampling. This combined optimization enables the VAE to proficiently tokenize and reconstruct SMILES while creating a structured latent space, laying the foundation for fine-tuning on toxicity-based constraints.

\subsection{Fine-Tuning on Target Properties}
\label{subsec:stage2}
Building upon the general-purpose latent space developed in Stage~1, we fine-tune the \emph{Pesti-Gen} to align with pesticide-specific toxicity metrics using dataset as described in Section~\ref{subsec:gyeonggi}. In particular, we incorporate both livestock and aqua ecotoxicity levels. Collectively, these provide a quantitative indicator of the environmental risks posed by the generated molecules.

\paragraph{Partial Freezing and Decoder Enhancement.}
To preserve the broad chemical representations learned previously, we fix the encoder parameters, denoted by \(\boldsymbol{\theta}_{e}\). Concretely, let
\[
\boldsymbol{\theta}_{e}^{*} = \arg\min_{\boldsymbol{\theta}_{e}} \mathcal{L}_{\text{pretrain}},
\]
where \(\mathcal{L}_{\text{pretrain}}\) is the original VAE objective (e.g., reconstruction plus KL divergence) minimized during the first training stage. These parameters \(\boldsymbol{\theta}_{e}^{*}\) remain frozen, thereby retaining the learned latent space structure. We then fine-tune the decoder with updated parameters \(\boldsymbol{\theta}_{d}'\), introducing new layers specifically handling the toxicity features:
\[
\boldsymbol{\theta}_{d}' = \arg \min_{\boldsymbol{\theta}_{d}} \Bigl(\mathcal{L}_{\text{recon}} + \alpha\,\mathcal{L}_{\text{toxicity}}^{(\text{livestock})} + \beta\,\mathcal{L}_{\text{toxicity}}^{(\text{aqua})}\Bigr).
\]
This formulation ensures that toxicity constraints for both \(\textit{livestock}\) and \emph{aqua}s impact are seamlessly incorporated into the decoding path, guiding the model to produce eco-friendly pesticide candidates without loosing its general chemical structure. By isolating the pre-trained encoder, the integrity of the original representations is preserved, allowing the subsequent fine-tuning process to focus on embedding toxicity considerations into the generative output.

\paragraph{Latent Noise Injection.}
To improve robustness and encourage smoother latent representations, we inject isotropic Gaussian noise into the latent embeddings coming from the frozen encoder. Let \(\mathbf{z}\) be the latent output of the encoder; we form a noisy vector:
\[
\mathbf{z}' = \mathbf{z} + \boldsymbol{\epsilon}, \quad 
\boldsymbol{\epsilon} \sim \mathcal{N}(0,\,\sigma^2 \mathbf{I}),
\]
where \(\sigma\) is the noise level and \(\mathbf{I}\) is the identity matrix. Training the decoder on these perturbed embeddings fosters stability under minor distortions, thereby reducing overfitting and improving the model’s ability to generalize to novel molecular structures.

\paragraph{Toxicity-Aware Loss Function.}
Finally, we augment the traditional VAE loss by adding separate penalty terms for \(\textit{LD}_{50}\) and ecological toxicity. Our revised objective is:
\[
\mathcal{L}' \;=\; \mathcal{L}_{\text{KL}} \;+\; \mathcal{L}_{\text{recon}}
\;+\;
\alpha\,\mathcal{L}_{\text{toxicity}}^{(\text{livestock})}
\;+\;
\beta\,\mathcal{L}_{\text{toxicity}}^{(\text{aqua})},
\]
where \(\mathcal{L}_{\text{KL}}\) is the standard KL divergence term, \(\mathcal{L}_{\text{recon}}\) measures SMILES reconstruction error (e.g., cross-entropy on token predictions), and \(\mathcal{L}_{\text{toxicity}}^{(\text{livestock})}\) and \(\mathcal{L}_{\text{toxicity}}^{(\text{aqua})}\) penalize deviations from desired \(\textit{livestock}\) and aqua toxicity levels, respectively. The hyperparameters \(\alpha\) and \(\beta\) govern the trade-offs among structural fidelity and ecological toxicity. Specifically, in our experiments, we set both \(\alpha = 0.5\) and \(\beta = 0.5\) for maintaining chemical viability while meeting stringent livestock toxicity and aqua eco-toxicity benchmarks.

\section{Results}

\begin{figure*}[t]
    \centering
    \includegraphics[width=1\textwidth]{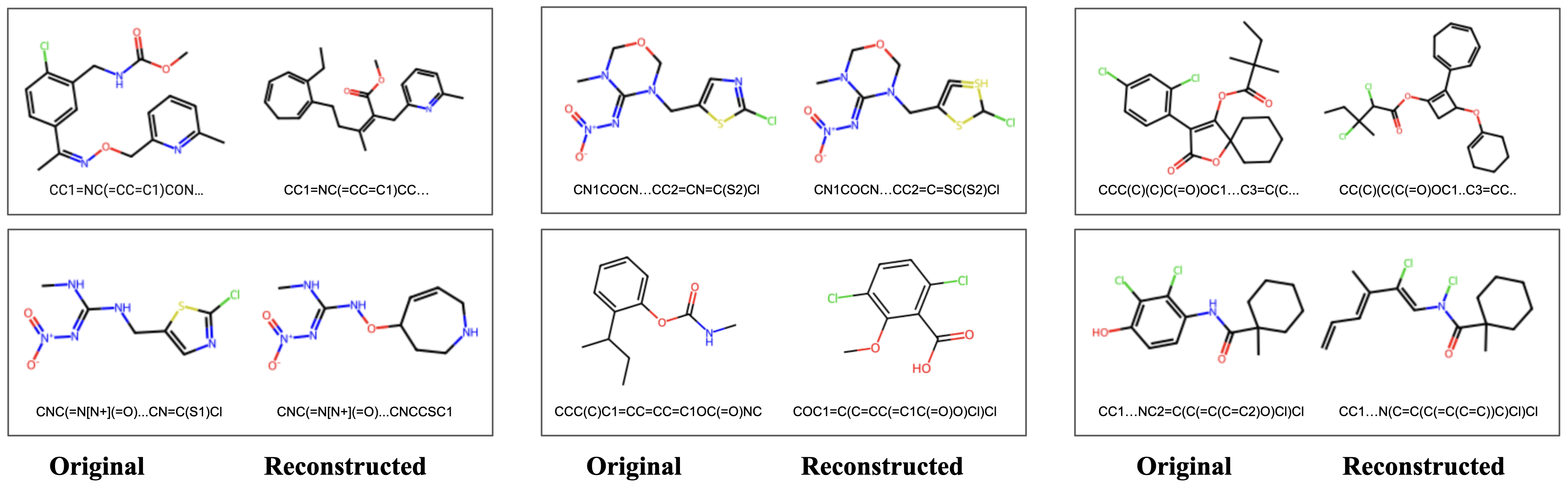}
    \caption{\textbf{Samples of reconstructed pesticide molecules generated by the trained model}. Each pair consists of an original molecule (left) and its corresponding reconstructed molecule (right).}
    \label{fig:smiles_validity}
\end{figure*}

In this section, we present the outcomes of our \emph{Pesti-Gen} model, focusing on SMILES validity, novelty, and key physicochemical characteristics. We first describe the filtering strategy that ensured the generation of chemically plausible candidates. Next, we analyze the distribution of LogP values, and also compare Synthetic Accessibility Scores (SAS). Finally, we discuss the structural analysis. 

\subsection{Candidate Validity and Novelty}
\label{subsec:valid_smiles}

We used cheminformatics tools such as RDKit\footnote{\url{https://www.rdkit.org/docs/Overview.html}} to validate the SMILES strings generated by our model. This validation step discards syntactically malformed or chemically implausible SMILES, retaining only those that represent legitimate molecular structures. This process is crucial for downstream analyses, as it ensures that subsequent evaluations of physicochemical and toxicity-related properties are conducted on structurally coherent and innovative candidates. 

Through iterative sampling and filtering, the model achieved a SMILES structure validity rate of approximately 68\%. This underscores the robustness of the learned latent representation and the feasibility of leveraging the outputs for further refinement. Additionally, the distribution of sequence similarities (measured using 1-Normalized Levenshtein Distance) fell within the range of 0.6 to 0.68, indicating that the generated molecules were both valid and structurally novel. 

\subsection{Additional Property Distribution Analysis}
\label{subsec:logp_analysis}

Table~\ref{table:Pesti-Gen-compact-final} summarizes the key findings regarding LogP (octanol-water partition coefficient) and distribution metrics for both original and \emph{Pesti-Gen}-generated molecules.
\begin{table}[h]
\centering
\begin{tabular}{lcc}
\hline
\textbf{Metric} & \textbf{Original} & \textbf{Generated} \\
\hline
LogP Range & [-6.75, 5.94] & [-3.90, 8.60] \\
Mean LogP & 2.81 & 2.89 \\
LogP SD & 1.98 & 2.14 \\
Validity (\%) & 100 & 67.86 \\
Mean Dist. Vector & [-0.39, 0.44] & [-0.20, 0.20] \\
\hline
\end{tabular}
\caption{Performance of Pesti-Gen in terms of LogP, validity rates, and mean distribution.}
\label{table:Pesti-Gen-compact-final}
\end{table}
Overall, the generated molecules exhibit a slightly broader LogP range, reflecting the model’s capacity to explore a wider chemical space. The mean LogP and its standard deviation are marginally higher for the generated compounds, indicating a shift toward more hydrophobic structures in some instances. While a higher LogP may not always align with the ideal physicochemical profile for every application, it demonstrates the model's ability to push the boundaries of chemical diversity. Moreover, the majority of generated molecules remain within a reasonable range, suggesting that this variation does not compromise the general utility of the candidates. 

\label{subsec:sas_analysis}

Table~\ref{table:Pesti-Gen-sas-comparison} compares the Synthetic Accessibility Scores (SAS) of the original and generated molecules, highlighting the model’s influence on synthetic feasibility. The generated set also shows a mildly lower mean SAS, suggesting a modest improvement in synthetic accessibility on average. A lower SAS score generally indicates that molecules are easier to synthesize, which is advantageous for practical applications like pesticide production. 

However, simplicity in structure does not always equate to ideal functionality, as some applications may require more complex designs. The narrower range of SAS scores for generated molecules highlights a more consistent synthetic feasibility, while the slightly increased standard deviation reflects a subset of molecules with higher synthetic complexity. Importantly, both the original and generated datasets share similar extremes in SAS values, indicating that the model retains boundary conditions of synthetic difficulty while expanding diversity.

\begin{table}[h]
\centering
\begin{tabular}{lcc}
\hline
\textbf{Metric} & \textbf{Original} & \textbf{Generated} \\
\hline
Mean SAS & 0.5985 & 0.5378 \\
SAS STD Dev & 0.2095 & 0.2375 \\
SAS Range & [0.1461, 0.8850] & [0.1313, 0.8344] \\
\hline
\end{tabular}
\caption{Comparison of Synthetic Accessibility Scores (SAS) between original and Pesti-Gen-generated SMILES.}
\label{table:Pesti-Gen-sas-comparison}
\end{table}

Taken together, these results demonstrate that \emph{Pesti-Gen} can generate structurally varied molecules with a balanced trade-off between hydrophobicity and synthetic accessibility. While higher LogP and lower SAS may not be optimal in all scenarios, they can still be beneficial depending on the specific requirements of pesticide design. Additionally, with more refined or targeted datasets, the outcomes are expected to align even better with desired characteristics. 


\subsection{Structural Analysis and Visualization}

Figure~\ref{fig:smiles_validity} illustrates the transformation of an original molecule into its reconstructed counterpart, showcasing the model’s ability to generate eco-friendly pesticide candidates while retaining key functional features. For instance, below shows one of the reconstructed molecule.

\textbf{Original:}  
\begin{equation}
    \footnotesize \texttt{CN1COCN(C1=NN+[O-])CC2=CN=C(S2)Cl}
\end{equation}

\textbf{Reconstructed:}  
\begin{equation}
    \footnotesize \texttt{CN1COCN(C1=NN+[O-])CC2=C=SC(S2)Cl}
\end{equation}

In this equation, the reconstructed molecule retains the \textit{nitro-substituted ring} (\(N[N+](=O)[O-]\)), which is critical for pesticidal efficacy, while introducing subtle adjustments to the \textit{sulfur-containing heterocycle} (\(C=SC(S2)\)). These modifications have the potential to improve biodegradability and reduce environmental persistence, aligning with the objective of designing novel ecofriendly pesticide candidates. At the same time, the core scaffold remains intact, ensuring preserved activity and functionality. These reconstructed molecules demonstrate that the model produces structurally plausible and functionally relevant molecules, emphasizing its feasibility as a tool for eco-friendly pesticide design.

Considering that this task optimizes over two toxicity scores—livestock toxicity and aqua toxicity—the ability to generate novel candidates for such a multi-objective problem highlights the model's feasibility and versatility. This demonstrates its potential to handle complex optimization tasks, suggesting that even greater improvements are achievable with specialization on a single toxicity metric. This example highlights the model’s capacity to balance structural innovation with ecological considerations, producing candidates that are both functional and potentially less harmful to the environment. By focusing on subtle but meaningful modifications, the generated molecules offer a promising starting point for generative model based sustainable pesticide development.

\section{Conclusion}
This study successfully demonstrated the feasibility of generating novel pesticide candidates by adapting pipelines originally designed for drug discovery, even when working with a limited dataset. If a specific target within the pesticide dataset or additional internal dataset is incorporated, the performance is expected to improve further, potentially yielding even better results. Additionally, the loss function can be tailored for specific target purposes—for instance, incorporating other toxicity conditioning or optimizing for alternative metrics—to better align with specialized objectives.

\section{Limitation and Future Works}
While this study establishes a foundational approach for eco-friendly pesticide generation and optimization, it also faces notable constraints. Reliance on publicly accessible, license-free datasets limits the model's capacity to learn nuanced pesticide-specific knowledge, and any computational predictions ultimately require wet-lab experimentation for validation. Additionally, although our multi-objective approach underscores the model’s versatility, more focused single-objective optimization could further refine toxicity profiles, and incorporating cross-resistance tracking could help agronomists devise effective rotation strategies. 

Nevertheless, the newly compiled dataset—comprising 56,360 data points and 520 unique pesticides—presents a significant resource for future research. Detailed geographic coordinates (latitude and longitude) enable environment-specific analysis of pesticide behavior, and basing the dataset on LC\textsubscript{50} and LD\textsubscript{50} metrics can yield more robust toxicity classifications and ecological impact assessments. As future work explores expanded datasets or more specialized objectives, these elements collectively offer a richer basis for enhanced pesticide design and sustainable agricultural practices.

\bibliography{main}
\end{document}